\documentclass[letterpaper, 10 pt, conference]{ieeeconf}  % Comment this line out if you need a4paper

\IEEEoverridecommandlockouts                              % This command is only needed if 
\usepackage[table]{xcolor} %% colored text and table row highlighting
\usepackage{balance} %% to make references end on the same line in a two-column setup
\usepackage{hyperref} %% to write urls
\usepackage{amsfonts,amsmath,amssymb}
\usepackage{booktabs}
\usepackage{graphicx}
\usepackage{array}
\usepackage{xspace}

\usepackage{amsthm}
\newtheorem*{remark}{Remark}

\usepackage{siunitx}
\usepackage{tabularx}
\usepackage[caption=false]{subfig} % Preserves IEEE caption style
\usepackage{mwe}
\usepackage[ruled,vlined,linesnumbered]{algorithm2e}
\usepackage{algpseudocode}
\usepackage{colortbl}

\definecolor{codegray}{gray}{1}
\definecolor{styellow}{RGB}{200,160,0}
\definecolor{rrtblue}{RGB}{40,80,200}
\definecolor{red}{RGB}{255,0,0}
\definecolor{green}{RGB}{0,255,0}
\definecolor{new}{RGB}{0,140,60}

\newcommand\benchmarkName{DynoFluxBench\xspace}

\author{Franz Queißner$^1$, Andreas Orthey$^1$, Wolfgang Hönig$^{1}$%
\thanks{$^{1}$TU Berlin, Germany}%
}

\title{\Huge DynoFluxBench: Benchmarking Kinodynamic Space-Time Planners in Dynamic Environments}

\begin{document}

\maketitle
\begin{abstract}
Robots that leave structured, static environments must plan motions that are kinodynamically feasible and safe among moving obstacles. 
However, there are no dedicated benchmark frameworks that combine both aspects.
To overcome this, we present DynoFluxBench, a framework to compare kinodynamic
  planners in known, dynamic environments with unbounded arrival time.
To demonstrate its utility and establish strong baselines, we develop three dedicated planners, named ST-Db-RRT, ST-GBRRT, and KIST, that fuse kinodynamic and space-time methods, covering different kinodynamic search paradigms: 
ST-Db-RRT expands with randomly selected discontinuity-bounded motion primitives using trajectory optimization, whereas KIST and ST-GBRRT maintain a kinodynamically feasible tree with different heuristic guidance. 
We analyze the probabilistic completeness guarantees of those new planners in dynamic environments. Finally, we evaluate ST-Db-RRT, ST-GBRRT, and KIST using DynoFluxBench, showing that ST-Db-RRT reaches a first solution up to 32 times faster, while KIST and ST-GBRRT remain valuable where trajectory optimization is fragile. 
Videos and further analysis can be found at~\url{https://dynofluxbench.github.io/dynofluxbench/}.
\end{abstract}
%This establishes DynoFluxBench as a comprehensive evaluation framework for kinodynamic planning in dynamic environments. 

\section{Introduction}
\begin{figure}[t]
  \centering
  \includegraphics[width=\linewidth]{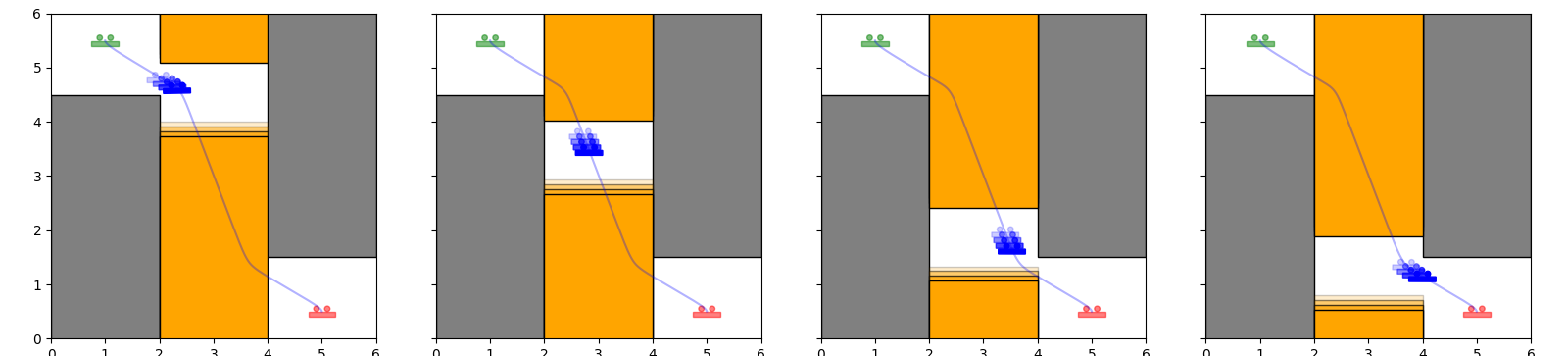}\\[0.35em]
  \includegraphics[width=\linewidth]{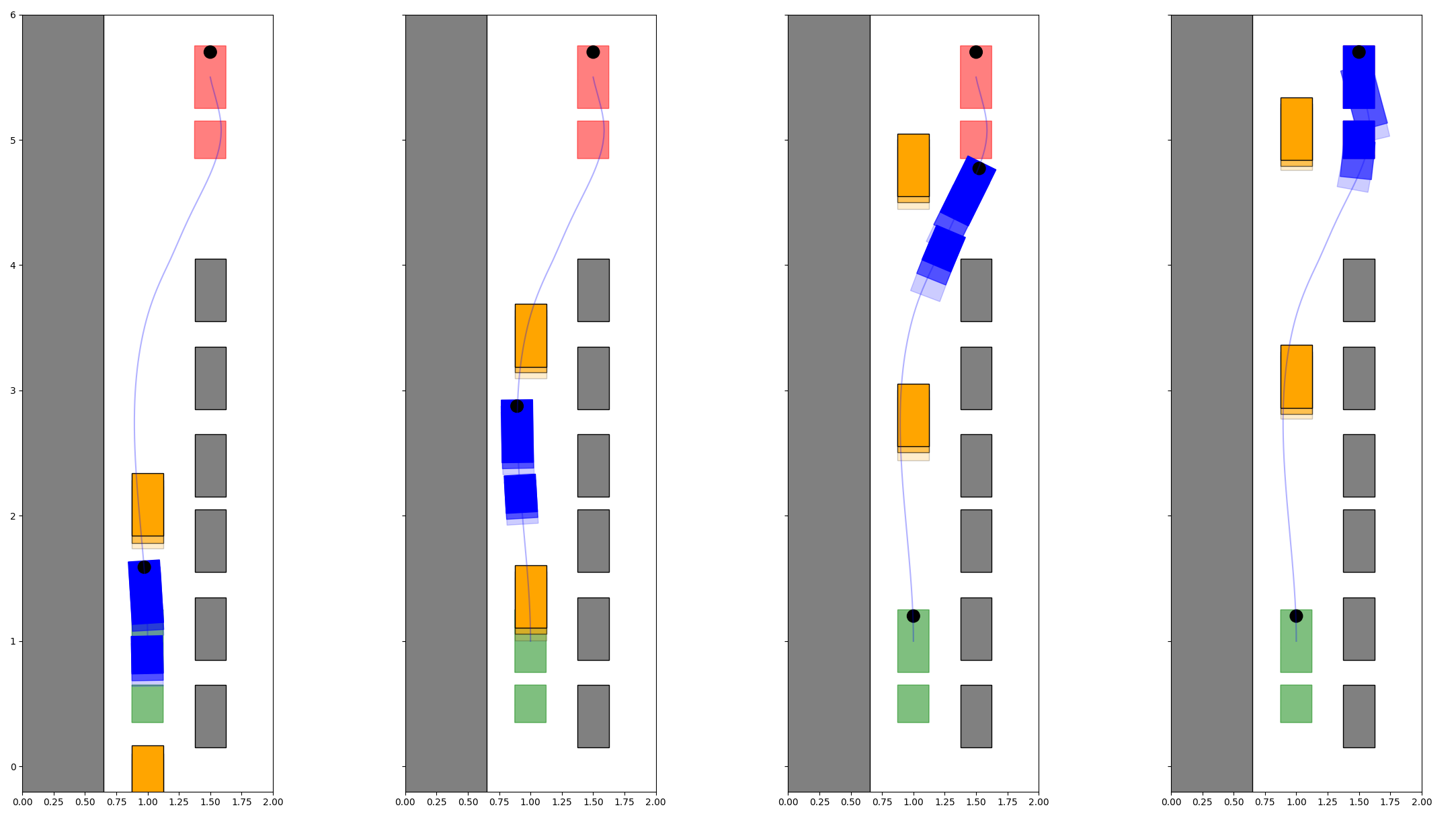}\\[0.35em]
  \includegraphics[width=\linewidth]{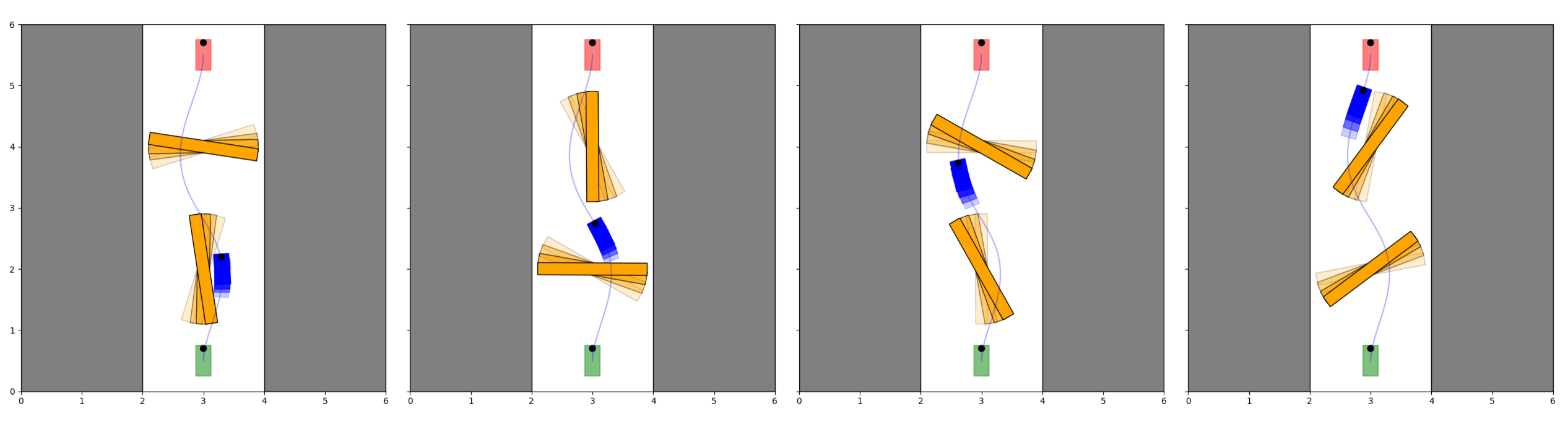}
  \caption{Example kinodynamic trajectories in dynamic environments from
    \benchmarkName.
    \textbf{Top:} Planar rotor timing its passage through a moving elevator
    gap.
    \textbf{Middle:} Car performing a lane switch among moving traffic.
    \textbf{Bottom:} Second-order unicycle traversing two revolving doors.
    Each row shows four snapshots along the planned space-time path
    (green: start, red: goal, orange: dynamic obstacles).}
  \label{fig:pullfigure}
\end{figure}

Planning feasible kinodynamic motions and avoiding dynamic obstacles have both been studied intensively, but largely in isolation. The first is addressed by \textit{kinodynamic} motion planning~\cite{rrt,Sucan2009BKPIECE,sst}, which takes the kinematic and differential constraints of the robot into account, such as the turning rate of a car or the limited acceleration of a torque-constrained system. The second is addressed by planning in \textit{known dynamic environments}~\cite{statetimespace,timebasedrrt,strrt}: when the trajectories of moving obstacles are known in advance, as, for example, in multi-robot planning, we can treat time as an additional dimension, and the planner can anticipate the motion of the obstacles~\cite{strrt}.

%Each subproblem is solved well by the current state of the art. 
Recent kinodynamic planners such as Iterative Discontinuity-bounded Rapidly-exploring Random Tree (iDb-RRT)~\cite{ortizharo2024idbrrtsamplingbasedkinodynamicmotion} and Generalized Bidirectional RRT (GBRRT)~\cite{gbrrt} find kinodynamically feasible trajectories for a wide range of systems, but only in static environments. By contrast, Space-Time-RRT* (ST-RRT*) \cite{strrt} plans optimally through space-time with moving obstacles and unknown arrival time, but only for holonomic or steering-capable systems. 
While kinodynamic planners for dynamic environments
exist~\cite{vdBerg2007KinodynamicDynamic,rrtx}, they require a steering function
and do not address problems with unbounded time. 
To the best of our knowledge, there exists no kinodynamic motion planner for general dynamical systems that simultaneously avoids steering functions and plans through space-time with dynamic obstacles and unknown, unbounded arrival time.
%Lifting planners like ST-RRT*~\cite{strrt} into the kinodynamic setting, however, is not straightforward, as it requires solving a two-point boundary value problem (BVP), which is computationally expensive or not available for many systems. The kinodynamic planners iDb-RRT and GBRRT, however, both avoid the steering function, by deferring feasibility to one single optimization step or by never leaving the set of feasible motions. 

To combine both aspects, we develop three new planners, ST-Db-RRT, ST-GBRRT, and KIST, that plan for kinodynamic systems in space-time for dynamic environments \emph{without requiring a steering function or a known arrival time}. All three planners share the same space-time search strategy but apply it to different kinodynamic search paradigms. Beyond the algorithms themselves, we analyze the theoretical consequences of the space-time extension, showing that probabilistic completeness of GBRRT is preserved, but the resolution completeness of Db-RRT is only preserved with bounded time. 
%The asymptotic optimality of ST-RRT* through rewiring is not preserved in the kinodynamic setting, though it remains a promising direction for future work.
Those three planners are used as baselines for \benchmarkName, a novel benchmark which evaluates kinodynamic space-time planners on a set of dynamic environments (Fig.~\ref{fig:pullfigure}).
To summarize, we make the following contributions:

\begin{itemize}
    \item \textbf{\benchmarkName:} Create a novel benchmark set to evaluate kinodynamic space-time planners on 25 scenarios involving four dynamical systems.
    \item \textbf{Three Planners:} Extension of existing kinodynamic planners iDb-RRT~\cite{ortizharo2024idbrrtsamplingbasedkinodynamicmotion} and GBRRT~\cite{gbrrt} to space-time obtaining ST-Db-RRT and ST-GBRRT, respectively. Additionally, we develop KIST, a combination of different aspects of ST-Db-RRT and ST-GBRRT. 
    %\item \textbf{Incompleteness Proof:} We show that a naive extension of Db-RRT~\cite{ortizharo2024idbrrtsamplingbasedkinodynamicmotion} is incomplete in dynamic environments.
    \item \textbf{Theoretical Analysis:} Analysis of the algorithmic
      properties of Naive Db-RRT, ST-Db-RRT, ST-GBRRT, and KIST pertaining to dynamic environments.
\end{itemize}

\section{Related Work}
\label{ch:relatedwork}

Motion planning in dynamic environments and kinodynamic motion planning are well
studied fields. In the following, we review both fields as well as the existing
works at their intersection.

\subsection{Planning in Dynamic Environments}
Planning in dynamic environments is often approached reactively, without assuming prior knowledge of the obstacle trajectories. Execution-extended RRT \cite{errt} reuses waypoints of previous solutions to speed up replanning, Real-Time RRT* \cite{rtrrt} interleaves online tree rewiring with path execution, and RRTX \cite{rrtx} maintains an asymptotically optimal search tree that is repaired online. 

When the obstacle trajectories are known in advance, the motion of obstacles can be anticipated. Path-velocity decomposition~\cite{pathvelocity} does so in two stages, planning a geometric path first and a velocity profile along it that avoids the moving obstacles. Committing to a fixed path early, however, can render every velocity profile infeasible. A more general approach is to treat time as an explicit planning dimension by planning in space-time~\cite{statetimespace}.
Time-Based RRT~\cite{timebasedrrt} leverages space-time to plan a rendezvous of two dynamic systems with known arrival time. SIPP~\cite{sipp} compresses the time dimension into safe intervals and computes optimal trajectories, but requires a discretization of the state space, restricting it to low-dimensional problems.
Sampling-based planners like T-PRM~\cite{tprm} avoid this state space discretization by augmenting the roadmap with a time dimension, while SI-RRT~\cite{sirrt} transfers the safe-interval representation into an RRT-Connect search for high-DoF manipulators, an idea also applied to multi-robot planning~\cite{sirrtstar}. Both, however, are restricted to holonomic systems.
ST-RRT*~\cite{strrt} lifts RRT* into space-time without assuming prior knowledge of the arrival time, planning asymptotically optimal trajectories in dynamic environments. However, its extension step interpolates linearly between states, restricting it to holonomic systems with velocity limits, and kinodynamic constraints cannot be handled.

\subsection{Kinodynamic Motion Planning}

Kinodynamic motion planning accounts for differential constraints and actuation limits. There are mainly three categories of approaches:
Search-based methods discretize the control space with kinodynamically feasible motion primitives \cite{statelattice} to construct a graph of reachable states, which can then be searched with search algorithms like A*. This approach provides strong theoretical guarantees relative to the chosen discretization, but the size of the graph grows exponentially with the state dimension, making exploration of high-dimensional state spaces infeasible.

Sampling-based planners explore the state space more rapidly by growing a tree~\cite{rrt} towards randomly sampled states using propagated controls, while preserving completeness under correct propagation~\cite{kleinbort2022probabilisticcompletenessrrtgeometric}. RRT methods can achieve asymptotic optimality by planning in state-cost space (AO-RRT) \cite{aorrt}, using witness sets (SST*) \cite{sst} or rewiring the tree (RRT*) \cite{rrtstar}, \cite{rrtx}, \cite{strrt}. Rewiring, however, requires connecting two states exactly which in the kinodynamic case can be described by a two-point boundary value problem (BVP) that can be solved in closed form only for restricted system classes such as linear dynamics \cite{kinodynamicrrtstar}, and is computationally expensive or unsolvable for general kinodynamic systems. Heuristics can be used to improve sampling with informed sets \cite{informedrrtstar} or AIT* \cite{aitstar}, which computes a cost-to-go heuristic through an improving reverse search tree. GBRRT~\cite{gbrrt} makes this feasible for the general kinodynamic setting by avoiding the requirement to solve any BVP using a heuristic-guided bidirectional search with forward propagation.

Optimization-based methods such as CHOMP \cite{chomp} or sequential convex optimization \cite{trajopt} optimize trajectories while incorporating complex constraints well. Their performance is highly dependent on an initial guess and suffers from local minima~\cite{Orthey2020RAL}. T-CHOMP \cite{tchomp} extends the CHOMP \cite{chomp} trajectory optimization framework to space-time, enabling it to handle time-dependent constraints, but requires a prior fixed trajectory duration.
More recently, hybrid methods like db-A* \cite{hoenig2022dbadiscontinuityboundedsearchkinodynamic,ortizharo2023idbaiterativesearchoptimization} connect precomputed motion primitives while tolerating a bounded discontinuity at the junctions, which is subsequently repaired by trajectory optimization to an optimal solution. iDb-RRT \cite{ortizharo2024idbrrtsamplingbasedkinodynamicmotion} embeds this discontinuity-bounded expansion into the RRT framework, removing the need to solve a BVP.
These planners efficiently solve kinodynamic problems in static environments. Time, however, is not an explicit planning dimension, and moving obstacles cannot be accounted for.

%%%%%%%%%%%%%%%%%%%%%%%%%%%%%%%%%%%%%%%%%%%
\subsection{Kinodynamic Planning in Dynamic Environments}
A smaller body of work addresses both challenges simultaneously. One of the earliest randomized kinodynamic planners~\cite{hsu2002randomized} builds a tree of milestones in state-time space, avoiding moving obstacles with known trajectories. While general with respect to the system dynamics, the planner expands only at random within a bounded time horizon, without goal-directed guidance and without any notion of solution quality. Chen et al. \cite{stsehs} extend the Space Exploration Guided Heuristic Search to the time domain, guiding a kinodynamic search with a time-dependent heuristic. The approach targets car-like robots and relies on a circle-based decomposition of the workspace. More recently, safe intervals~\cite{sipp} have been combined with kinodynamic models~\cite{sippkinodynamic,sippquad}, which retain completeness and optimality. 
However, each of those works is either restricted to a specific class of systems or depends on a discretization of the state space.

In summary, the reviewed planners for dynamic environments handle moving obstacles but assume holonomic systems, kinodynamic planners handle differential constraints but assume static environments, and the few works at their intersection are tied to specific system classes or state-space discretizations. We address this integrated problem with sampling-based planners that combine kinodynamic planning in dynamic environments with an unknown arrival time. %This thesis builds upon three algorithms that mark the state of the art of the respective subproblems: ST-RRT* \cite{strrt} contributes the space-time search for unbounded time, iDb-RRT \cite{ortizharo2024idbrrtsamplingbasedkinodynamicmotion} contributes fast kinodynamic exploration and GBRRT \cite{gbrrt} contributes an optimization-free approach. 

\section{Problem Statement}

Let $\mathcal{X} =
\mathcal{Q} \times \mathcal{T}$ be a space-time state space~\cite{strrt}, which pairs the configuration state space $\mathcal{Q}$ with the time axis $\mathcal{T}$. A space-time state $x = (q,\, t) \in \mathcal{X}$ therefore consists of a configuration state $q \in \mathcal{Q}$ and a time $t \in \mathcal{T}$. The system dynamics act on the configuration component only: applying a control $u \in \mathcal{U}$ for one timestep advances it by $q_{k+1} = step(q_k,\, u_k)$ following an Euler integration with a known system velocity limit $\dot{q}_{\mathrm{max}}$, while the time component advances deterministically by the fixed step $\Delta t$. The resulting space-time transition is
\begin{equation}
  \label{eq:sttransition}
  x_{k+1} = (q_{k+1},\, t_{k+1}) = \bigl(\,step(q_k,\, u_k),\; t_k + \Delta t\,\bigr).
\end{equation}
Let $\mathcal{X}_{\mathrm{free}}(k) \subseteq \mathcal{X}$ be the collision-free
states at time $k$, $x_{\mathrm{start}}$ the start state, and
$\mathcal{X}_{\mathrm{goal}}$ the goal region. 
%We set the start time $t_{\mathrm{start}}=0$.
The lower-bound time to cover the distance between two configurations is denoted $\underline{t}(q_1,\, q_2)$. The desired output is a solution path $\mathcal{S}=(X,U)$ connecting $x_{\mathrm{start}}$ to the goal region $\mathcal{X}_{\mathrm{goal}}$ with a state sequence $X = [x_{\mathrm{start}}, x_1,\ldots,x_{\mathrm{goal}}]$, $x_{\mathrm{goal}} \in \mathcal{X}_{\mathrm{goal}}$, of length $K$ and a control sequence $U=[u_0, u_1,\ldots,u_{K-1}]$ such that $x_k \in \mathcal{X}_{\mathrm{free}}(k)$. The cost of a feasible path is $c = K \cdot \Delta t$.
%For KIST and ST-GBRRT the solution is kinodynamically feasible, $x_{k+1} = step(x_k, u_k)$ $\forall x_k \in X$; for ST-Db-RRT it is only \textit{discontinuity-bounded}~\cite{ortizharo2024idbrrtsamplingbasedkinodynamicmotion}.
%Since the cost is solely the trajectory duration, the cost of a feasible path is $c = K \cdot \Delta t$. For the discontinuity-bounded solution, node times follow Eq.~\ref{timeintegration}, which additionally accounts for the lower-bound time at each junction.

\section{Kinodynamic Space-Time Primitives\label{sec:motionprimitives}}

To plan in kinodynamic space-time, we use goal region expansion and distance computation from space-time planning~\cite{strrt} and discontinuity-bounded expansion with motion primitives for kinodynamic systems~\cite{ortizharo2024idbrrtsamplingbasedkinodynamicmotion}.

%All three planners presented in this chapter share the same \emph{space-time search strategy}, which they adopt from ST-RRT*~\cite{strrt} which enables them to plan in dynamic environments. 
%Here we fix the terminology and clarify the implementation details that the planners have in common. For a detailed pseudocode of the shared functions we refer to Algorithms 2-5 in the ST-RRT* paper \cite{strrt}.

\subsection{Goal Region Expansion}

To handle unbounded space-time, we leverage the conditional sampling and the goal region expansion methods from ST-RRT*~\cite{strrt}. 

Conditional sampling samples a random configuration and then computes the lower-bound time to the start state, conditioning the sampled time on the reachability from the start. The upper time bound is defined by the last valid time the system could still reach the latest goal arrival time of all currently existing goals.

The goal region expansion uses the two parameters \textit{initial batch size} and \textit{range factor} to progressively update the goal region. 
In every iteration, we check if the current sampling batch is exhausted and if so, expand the goal region by updating the time bounds by the value of the \textit{range factor}. Thus, the initial batch size controls how many samples are drawn in the time bounds before the goal region is expanded, and the range factor controls how much the goal region is expanded. 
When the planners find a first solution, the progressive goal region expansion is stopped. Instead, the upper time bound is set to the arrival time of the best current solution, and the lower time bound is set to the lower-bound time from start to goal. In this way, only samples that could improve the current best solution are drawn.

\subsection{Distance Computation}

The distance between states is defined by the space-time
pseudometric~\cite{strrt}. This pseudometric consists of the intrinsic configuration distance $d_\mathcal{Q}$ and the temporal difference $d_\mathcal{T}$, which is weighted by $\lambda \in (0,1)$. For two given states, \emph{time-consistency} is enforced, which means
a state may only be reached from an earlier state, and only if the velocity required to cover the configuration gap within the available time respects the limits $\dot{q}_{\mathrm{max}}$; states violating either condition are infinitely far apart.
%
%\begin{equation}
%  \label{eq:timeconsistent}
%  \begin{aligned}
%  &\textsc{TimeConsistent}(x_n,\, x,\, T) \;\Longleftrightarrow\; \\
%  &\qquad
%  \begin{cases}
%    \underline{t}(q_n,\, q) \le t - t_n & \text{if } T = T_{\mathrm{fwd}},\\[2pt]
%    \underline{t}(q,\, q_n) \le t_n - t & \text{if } T = T_{\mathrm{rev}}.
%  \end{cases}
%  \end{aligned}
%\end{equation}
%

%We resolve this the same way in every planner: a small neighbourhood is queried with $d_\mathcal{Q}$ and the first candidate satisfying the time-consistency requirement (Eq.~\ref{eq:timeconsistent}) is returned (Alg.~\ref{alg:nearest}).

%\paragraph{Time-Consistency} %All nearest-neighbor queries use the same time-consistency function, which ensures both conditions of the space-time pseudometric (Eq.~\ref{distance-st}): 
%a tree node $x_n \in T$ is time-consistent with a query state $x$ if the configuration gap between them can be traversed within their time difference,

%Since $\underline{t} \ge 0$, this condition enforces the correct temporal ordering: a forward-tree node must lie earlier in time than the query state, and a reverse-tree node later. The time-consistent nearest-neighbour function (Alg.~\ref{alg:nearest}) queries a small neighbourhood in the k-d tree and returns the first candidate satisfying Eq.~\ref{eq:timeconsistent}.

\subsection{Kinodynamic Expansion Modes}
We leverage motion primitives to discretise the action space for fast kinodynamic expansion \cite{hoenig2022dbadiscontinuityboundedsearchkinodynamic}. A primitive is defined as $m = (X, U)$ consisting of a kinodynamically feasible sequence of states $X = [x_0, x_1,\ldots,x_f]$ and actions $U=[u_0, u_1,\ldots,u_{f-1}]$. 
Discretising the action space into a finite set of primitives means that, in general, no primitive's start state $x_0$ exactly matches the state being expanded, which introduces a discontinuity between that state and the primitive. Allowing a discontinuity up to a certain bound $\delta$ makes a trajectory \textit{$\delta$-discontinuity-bounded}. We use $x_f = x \oplus m$ to indicate that the motion $m$ is applied from a state $x$ with a \textit{$\delta$-discontinuity-bound}, landing in a final state $x_f$.

To expand and connect states with primitives, we use one of three selection policies:
\begin{equation*}
    \sigma \in \{\textsc{Rand},\; \textsc{Best},\; \textsc{Prop}\}.
\end{equation*}
With $\sigma = \textsc{Rand}$, the first valid, randomly drawn primitive is applied as a $\delta$-discontinuity-bounded jump $x \oplus m$. With $\sigma = \textsc{Best}$, the primitives are ordered by the distance of their landing state to the expansion target before the first valid one is applied, again as a discontinuity-bounded jump. With $\sigma = \textsc{Prop}$, the primitives are ordered as in \textsc{Best}, but the actions of a candidate are forward propagated from the parent state with the system dynamics, resulting in a kinodynamically feasible motion without discontinuity. 

%\paragraph{Random Rollout} 
%It randomly draws a control $u \in \mathcal{U}$ and propagates it from the node for a random duration $t \in [0, t_{\mathrm{max}}]$. 

%\paragraph{Motion Selection} \textsc{SelectMotion} (Alg.~\ref{alg:selectmotion}) implements all three selection modes in one function: the applicable primitives $\mathcal{L}$ are shuffled ($\sigma = \textsc{Rand}$) or sorted by the distance of their landing state to the target ($\sigma \in \{\textsc{Best}, \textsc{Prop}\}$), and then applied one after another until the first valid new state is found. An expansion is valid if every transformed state of the motion is collision-free at its time, while a rollout (Alg.~\ref{alg:action_rollout}) checks validity at every integration step.

By using the three selection policies, we can define both expansion and connection methods. 

\paragraph{Expand} The \textsc{Expand} method grows a tree $T$ towards a sampled state $x_{\mathrm{rand}}$. It queries the $k = \log(|T|+1)$ nearest neighbors, and calls the selection policy for the actual expansion. The first successful new state is added to the tree and returned. If the neighborhood is exhausted, the expansion fails. 

\paragraph{Connect} A connection attempts to greedily establish a connection between two states, by repeatedly expanding towards the target state using the selection policy, as long as the distance to the target decreases, checking every intermediate node.

%\input{src/algorithms/select-motion.tex}
%\input{src/algorithms/connect.tex}

%Every new node keeps track of its time: a rollout advances it with the space-time transition (Eq.~\ref{eq:sttransition}), while a discontinuity-bounded jump integrates the motion duration and the lower-bound time of discontinuity jump (Eq.~\ref{timeintegration}). A more detailed explanation will follow in Sec.~\ref{sec:naivdbrrt}.

%\paragraph{Connect} %The \textsc{isConnected} function (Alg.~\ref{alg:connected}) checks whether a node $x$ in one tree forms a connection to the \emph{other} tree. It queries the other tree $T_b$ within the discontinuity bound $\delta$ and returns the first time-consistent pair. Requiring Eq.~\ref{eq:timeconsistent} at the connection also guarantees that the gap between the two nodes can be traversed under ideal conditions.\\
\section{Kinodynamic Space-Time Planners}

By relying on the kinodynamic space-time primitives, we develop three novel planners by generalizing Db-RRT~\cite{ortizharo2024idbrrtsamplingbasedkinodynamicmotion} and GBRRT~\cite{gbrrt}. To motivate this, we first discuss a naive Db-RRT planner.

\subsection{Naive Db-RRT for Dynamic Environments\label{sec:naivdbrrt}}

A first naive approach to elevate Db-RRT to deal with dynamic environments is to keep the original state space (without an explicit time dimension) and add dynamic collision checks to avoid moving obstacles. Since we know the duration of every motion primitive we can keep track of each node's time. This can be achieved by using the time as cost which makes the cost-to-come represent the time of each node. The durations of the discontinuities between the motion primitives are accounted for using the lower time bound function $\underline{t}(q_1,\, q_2)$ \cite{ortizharo2023idbaiterativesearchoptimization}.
%A new nodes time $t'$ can be calculated with
%\begin{equation}
%    \label{timeintegration}
%    t' = t + m.t + \underline{t}(q,\, m.q_s) 
%\end{equation} where $t$ is the time of its predecessor, $m.t$ the duration of the motion primitive and $\underline{t}(q,\, m.q_s)$ the lower-bound time estimate the systems needs to get from $q$ to the start state of the motion $m.q_s$.
Knowing each node time enables time-dependent collision checks and allows the planner to find discontinuity-bounded solutions. 
As we show in Sec.~\ref{sec:theoretical-analysis}, Naive Db-RRT is not probabilistically complete (PC).
%in environments with moving obstacles. In order to repair the discontinuity-bounded solution we also need to include dynamic collision checking into the trajectory optimization which is described further in Sec.~\ref{sec:optimization}.

%%%%%%%%%%%%%%%%%%%%%%%%%%%%%%%%%%%%%%%%%%%%%%%%%%%%%%%%%%%%%%%%%%%%%%%%%%%%%%%%
\subsection{Planner 1: ST-Db-RRT\label{sec:stdbrrt}}
%%%%%%%%%%%%%%%%%%%%%%%%%%%%%%%%%%%%%%%%%%%%%%%%%%%%%%%%%%%%%%%%%%%%%%%%%%%%%%%%
%We present  an extension of Db-RRT \cite{ortizharo2024idbrrtsamplingbasedkinodynamicmotion} to space-time by using the bidirectional search approach of ST-RRT*~\cite{strrt}.

To enable planning in space-time with Db-RRT, we develop the \textbf{S}pace-\textbf{T}ime \textbf{D}iscontinuity-\textbf{b}ounded RRT (ST-Db-RRT). ST-Db-RRT is a bidirectional tree planner, which uses the same sampling mechanism as ST-RRT*~\cite{strrt} based upon sampling in space-time and adaptive goal expansion. Contrary to ST-RRT*, it uses the fast randomized expansion mechanism $\sigma = \textsc{Rand}$ of Db-RRT~\cite{ortizharo2024idbrrtsamplingbasedkinodynamicmotion} to expand both trees in space-time. To connect the two trees, the $\sigma = \textsc{Best}$ connection mode is used as discussed in Sec.~\ref{sec:motionprimitives}. Once a connection is found, the discontinuity-bounded solution is extracted.

Repairing the discontinuity-bounded solution in dynamic environments requires a time-dependent trajectory optimization. We adopt the Differential Dynamic Programming (DDP) repair of Db-RRT to achieve this.
%(\cite{ortizharo2024idbrrtsamplingbasedkinodynamicmotion}, Chapter 4 E), chosen for its fast convergence and for operating on a trajectory of fixed length and fixed $\Delta t$. This fixed time grid makes the extension to dynamic environments straightforward as neither the number of stages nor $\Delta t$ changes during optimization, the time at each state stays constant and known, so the obstacle configuration at that time can be determined directly. 
DDP solves trajectory optimization problems of the form
\begin{subequations}
  \label{eq:opt} 
  \begin{align}
    \min_{Q, U} &\sum_{k=0}^{K-1} c(q_k, u_k) + c_K(q_K)\label{eq:opt_cost} \\
    \text{s.t.}\;\;\; &q_{k+1} = step(q_k, u_k) \; \forall k \in \{0, \ldots, K-1\} \\
    &q_0 = q_{\mathrm{start}},
  \end{align}
\end{subequations}
and enforces constraints such as collision avoidance through a cost term evaluated at every iteration. We keep time separate from the state: it is not a decision variable but is passed to the cost function (Eq.~\ref{eq:opt_cost}) solely to evaluate the collision constraint. The algorithm terminates if a planner termination condition is fulfilled.

%In static environments, this collision cost was computed from a static signed distance function. In dynamic environments, the known time of each state is used to update every obstacle to its configuration at that time before the signed distance is computed. 

%%%%%%%%%%%%%%%%%%%%%%%%%%%%%%%%%%%%%%%%%%%%%%%%%%%%%%%%%%%%%%%%%%%%%%%%%%%%%%%%
\subsection{Planner 2: ST-GBRRT\label{sec:stgbrrt}}
%%%%%%%%%%%%%%%%%%%%%%%%%%%%%%%%%%%%%%%%%%%%%%%%%%%%%%%%%%%%%%%%%%%%%%%%%%%%%%%%
%\input{src/algorithms/stgbrrt.tex}

ST-GBRRT is a direct transfer of GBRRT~\cite{gbrrt} to dynamic environments. 
GBRRT introduced the idea of guiding a feasible forward tree towards the goal
with a cost-to-goal heuristic taken from a reverse tree, so that the two trees
never have to be connected and no boundary value problem needs to be solved. 

%We integrate the shared space-time search strategy (Sec.~\ref{meth:strrt}), and we grow 
We adapted GBRRT by growing the reverse tree with fast $\delta$-discontinuity-bounded motion primitives instead of GBRRT's reverse propagation.
The forward tree keeps GBRRT's feasible action rollouts, 
interleaving exploration of the state space with exploitation of the reverse-tree heuristic. Because the forward tree is kinodynamically feasible and grows directly into the goal region, ST-GBRRT returns a solution without any completion policy or trajectory optimization.
\subsection{Planner 3: KIST\label{sec:kist}}
%%%%%%%%%%%%%%%%%%%%%%%%%%%%%%%%%%%%%%%%%%%%%%%%%%%%%%%%%%%%%%%%%%%%%%%%%%%%%%%%
%\input{src/algorithms/kist.tex}
To combine aspects of both planners, we develop \textbf{KIST}, a \textbf{KI}nodynamic \textbf{S}pace-\textbf{T}ime planning algorithm for dynamic environments. Similar to ST-Db-RRT, sampling follows ST-RRT*~\cite{strrt} and expansion follows the Db-RRT~\cite{ortizharo2024idbrrtsamplingbasedkinodynamicmotion} method for both trees. However, similar to GBRRT~\cite{gbrrt}, no connection between the trees is attempted. Instead, the trees are considered to be connected when they are in $\delta$ proximity, similar to Db-RRT. Instead of expanding with discontinuities, KIST avoids optimization by growing a feasible forward tree with primitive action rollouts. Additionally, it has a probability to perform a random action rollout that propagates a random action from $\mathcal{U}$ for a random duration $t \in [0, t_{\mathrm{max}}]$. The reverse tree grows with $\delta$-discontinuity-bounded motion primitives, but after an approximate connection with the forward tree, the algorithm uses the path of the reverse tree solely as a guiding heuristic to expand the forward tree greedily to the goal. We use a greedy A*-like search. If this completion policy succeeds, the solution path is extracted.

\section{Theoretical Analysis\label{sec:theoretical-analysis}}
\begin{table}[t]
\caption[Motion Planners Overview]{Overview of the presented planners and properties.
The expansion and connection modes (\textsc{Rand}: random primitive; \textsc{Best}: best primitive; \textsc{Prop}: forward propagation of the best primitive);
% are introduced in Sec.~\ref{meth:toolbox}, \textsc{Prop}/\textsc{Best} denotes forward/reverse tree.
ST: plans in space-time;
TO: uses trajectory optimization;
RC: resolution complete with respect to the provided motion primitives;
RC*: resolution complete with respect to the provided motion primitives, but not complete in unbounded time;
PC: probabilistically complete;
%AO: asymptotically optimal.
%Db-RRT does not consider moving obstacles (--). Naive Db-RRT loses completeness in dynamic environments through duplicate pruning (Remark 1, Sec.~\ref{sec:naivdbrrt}). ST-Db-RRT is only resolution complete if using randomly propagated (not time-optimal) motion primitives. KIST and ST-GBRRT retain probabilistic completeness through random action rollouts. None of the planners is asymptotically optimal, as all target fast initial solutions.}
}
\centering
\scriptsize
\setlength{\tabcolsep}{2.5pt}
\begin{tabularx}{\linewidth}{@{}lcccc>{\raggedright\arraybackslash}X@{}}
\toprule
 & & & \multicolumn{2}{c}{Completeness} & \\
\cmidrule(lr){4-5}
Planner & ST & TO & static & dynamic & Key features \\
\midrule
%Db-RRT \cite{ortizharo2024idbrrtsamplingbasedkinodynamicmotion} & $\times$ & \checkmark & RC & -- & $\times$ & \textsc{Rand} db-expansion, bidirectional $\delta$-connect \\
%\addlinespace
Naive Db-RRT & $\times$ & \checkmark & RC & $\times$ & time-annotated nodes, time-dependent collision checks \\
\addlinespace
ST-Db-RRT & \checkmark & \checkmark & RC & RC* &  \textsc{Rand} db-expansion, \textsc{Best} $\delta$-connect in space-time \\
\addlinespace
KIST & \checkmark & $\times$ & PC & PC & \textsc{Prop}/\textsc{Best} expansion, completion policy after tree connection \\
\addlinespace
ST-GBRRT & \checkmark & $\times$ & PC & PC & \textsc{Prop}/\textsc{Best} expansion, continuous reverse-tree heuristic \\
\bottomrule
\end{tabularx}

\label{tab:planner-overview}
\end{table}

We next analyze the different properties of the algorithms. A complete overview is found in Tab.~\ref{tab:planner-overview}. We show the key features of each planner, along with a description of whether planning is done in space-time (ST), if trajectory optimization (TO) is used, and the respective completeness guarantees in static and dynamic environments. 
In this section, we provide the details on the probabilistic completeness (PC) guarantees in dynamic environments.

\newcommand{\subtitle}[1]{\noindent\textbf{#1}:}

%%%%%%%%%%%%%%%%%%%%%%%%%%%%%%%%%%%%%%%%%%%%%%%%
\subtitle{Naive Db-RRT}
%%%%%%%%%%%%%%%%%%%%%%%%%%%%%%%%%%%%%%%%%%%%%%%%
To start this off, we first state a negative result of PC for Naive Db-RRT. 
\begin{remark} Naive Db-RRT is not PC in dynamic environments.
\end{remark}
\begin{proof}[Proof Sketch]
%Let us show this by contradiction. 
Duplicate pruning (\cite{ortizharo2024idbrrtsamplingbasedkinodynamicmotion} Alg. 2, Line 14) of Db-RRT rejects a new node when a node would be placed in a $\delta$-neighborhood, i.e., it suppresses nodes too close in state space. Consider a configuration $q$ that must be occupied at time $t_1$ to pass through a door within a short temporal window. If one branch reaches close to $q$ before $t_1$, duplicate pruning can suppress a later branch that would reach $q$ at $t_1$ and thus will never find a solution.
Without planning in space-time, the requirement ``be at $q$, but at a later time'' cannot be represented. In space-time, $(q, t_0)$ and $(q, t_1)$ are different states and duplicate pruning can distinguish between them, as the time is included in the distance metric. 
\end{proof}

%%%%%%%%%%%%%%%%%%%%%%%%%%%%%%%%%%%%%%%%%%%%%%%%
\subtitle{ST-Db-RRT\label{sec:stdbrrt_theory}}
%%%%%%%%%%%%%%%%%%%%%%%%%%%%%%%%%%%%%%%%%%%%%%%%
ST-Db-RRT does not suffer from the duplicate pruning problem that Db-RRT has in dynamic environments, as the pseudometric takes time into account. Planning in space-time, however, has other theoretical implications on PC.
The proof of PC for iDb-RRT and the family of Db-Algorithms~\cite{hoenig2022dbadiscontinuityboundedsearchkinodynamic,ortizharo2023idbaiterativesearchoptimization,ortizharo2024idbrrtsamplingbasedkinodynamicmotion} uses the chain of balls argument~\cite{kleinbort2022probabilisticcompletenessrrtgeometric}. 
A core requirement for the proof is a positive probability of reaching the next ball in the chain, which they show for random forward propagation in Lemma 3.
The Db-Algorithms instead discretize the action space with motion primitives. Their proof iteratively adds motion primitives and decreases the discontinuity bound, so that eventually all motions required to reach the successor ball are considered. This assumes that adding motion primitives covers the state space asymptotically (see Def. 4-5 and Remark 3 in \cite{ortizharo2023idbaiterativesearchoptimization}). Their generation achieves this by randomly sampling start and goal states and using trajectory optimization, resulting in time-optimal motions.
This is the point where the proof breaks in space-time, since covering space-time also requires time-suboptimal motions.
The completeness argument therefore needs the primitive-generating method to place positive measure on time-suboptimal motions of varied duration. Random forward propagation with randomized duration satisfies this, achieving resolution completeness with respect to the generated motion primitives.

However, even with a primitive-generating method that covers space-time asymptotically, ST-Db-RRT is not complete, because the unbounded time dimension keeps it from reporting that no solution exists. Db-RRT can terminate and move to the next iteration of iDb-RRT with more motion primitives. ST-Db-RRT, on the other hand, never terminates and keeps searching the unbounded time dimension. A known or estimated upper bound on time could fix this.
%%%%%%%%%%%%%%%%%%%%%%%%%%%%%%%%%%%%%%%%%%%%%%%%
\subtitle{ST-GBRRT}
%%%%%%%%%%%%%%%%%%%%%%%%%%%%%%%%%%%%%%%%%%%%%%%%
ST-GBRRT is a direct transfer of GBRRT~\cite{gbrrt} to space-time, so we can adapt the original proof~\cite{gbrrt}. While the best-input selection of motion primitives is not complete, completeness is restored by the random action rollout. The space-time sampling strategy ensures that the goal region is expanded and that there is a positive probability of sampling a feasible goal arrival time. This ensures that ST-GBRRT retains PC in dynamic environments.

%%%%%%%%%%%%%%%%%%%%%%%%%%%%%%%%%%%%%%%%%%%%%%%%
\subtitle{KIST}
%%%%%%%%%%%%%%%%%%%%%%%%%%%%%%%%%%%%%%%%%%%%%%%%
KIST resembles the structure of ST-RRT* at a high level, but uses a mixture of motion primitives and action rollouts for expansion. Its PC follows ST-RRT*'s proof sketch and the same chain of balls argument~\cite{kleinbort2022probabilisticcompletenessrrtgeometric} as ST-Db-RRT (Sec.~\ref{sec:stdbrrt_theory}). As there, the best-input selection policy alone would lose completeness, since the limited set of primitives gives no guarantee of expanding into the next ball in space-time. We restore PC by expanding the forward tree with a random action rollout.
Finally, in the case of unbounded goal time, we follow the argument of ST-RRT*: through progressive goal region expansion and conditional sampling, there is a positive probability of eventually sampling a feasible goal arrival time and any node that is part of a feasible solution.

\begin{figure*}[t]
\centering
\captionsetup[subfloat]{font=scriptsize,farskip=1pt,captionskip=1pt}%

% Row 1
\begin{minipage}[t]{0.13\linewidth}\centering
  \subfloat[Revolving Door]{\fbox{\includegraphics[width=\linewidth]{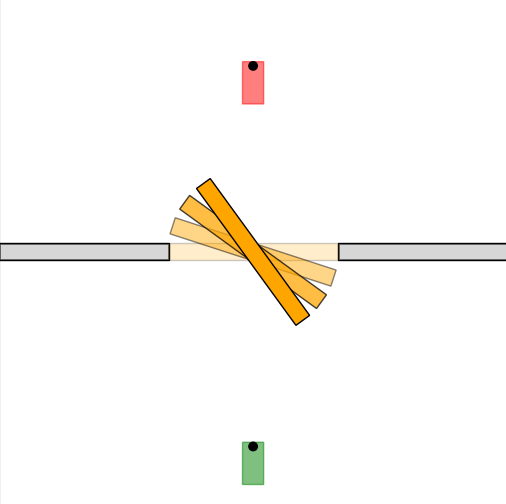}}}%
\end{minipage}\hfill
\begin{minipage}[t]{0.13\linewidth}\centering
  \subfloat[Double Revolving Door]{\fbox{\includegraphics[width=\linewidth]{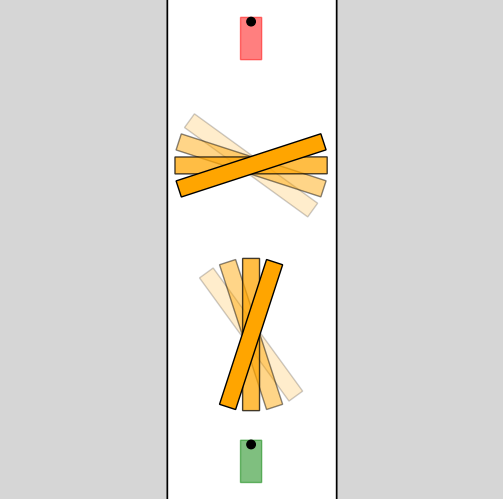}}}%
\end{minipage}\hfill
\begin{minipage}[t]{0.13\linewidth}\centering
  \subfloat[Multi Revolving Door]{\fbox{\includegraphics[width=\linewidth]{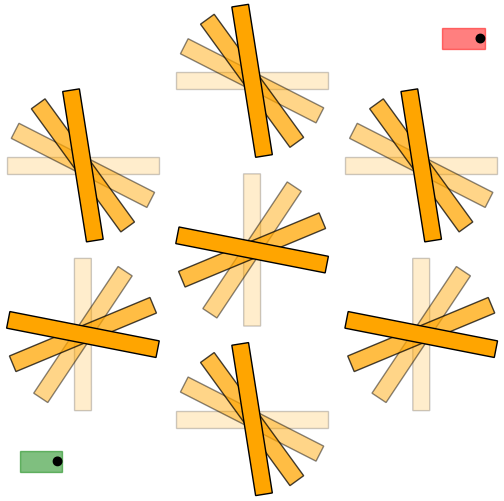}}}%
\end{minipage}\hfill
\begin{minipage}[t]{0.13\linewidth}\centering
  \subfloat[Double Moving Windows]{\fbox{\includegraphics[width=\linewidth]{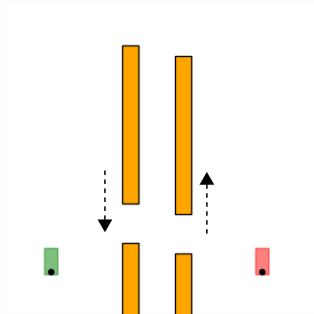}}}%
\end{minipage}\hfill
\begin{minipage}[t]{0.13\linewidth}\centering
  \subfloat[Shortcut]{\fbox{\includegraphics[width=\linewidth]{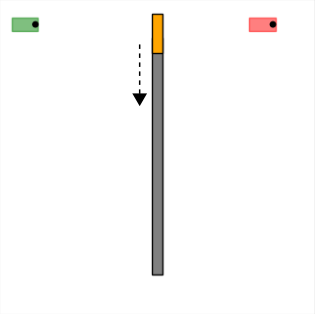}}}%
\end{minipage}\hfill
\begin{minipage}[t]{0.13\linewidth}\centering
  \subfloat[Bugtrap]{\fbox{\includegraphics[width=\linewidth]{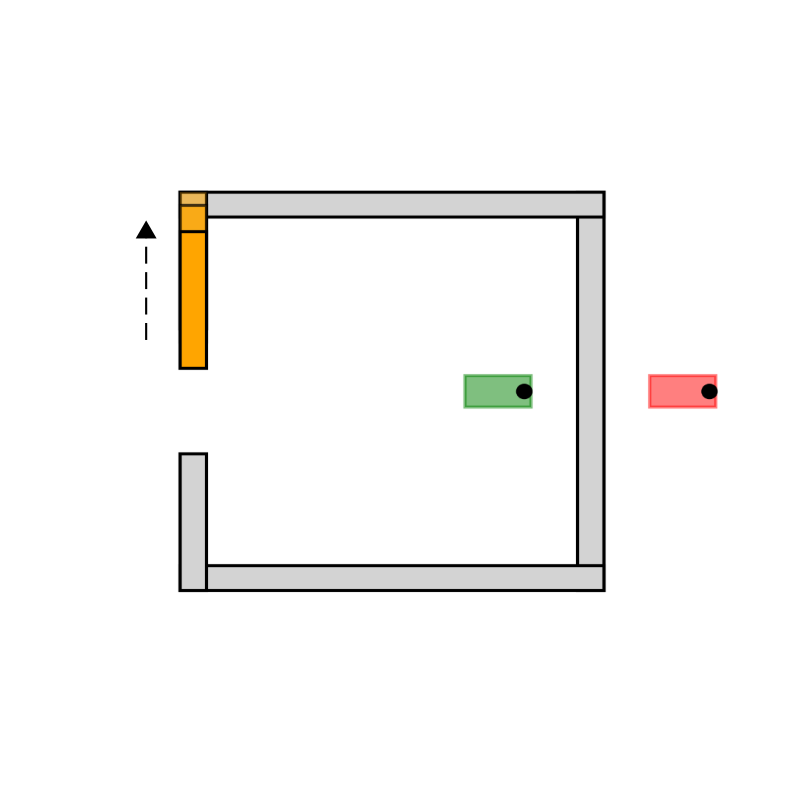}}}%
\end{minipage}\hfill
\begin{minipage}[t]{0.13\linewidth}\centering
  \subfloat[Wait]{\fbox{\includegraphics[width=\linewidth]{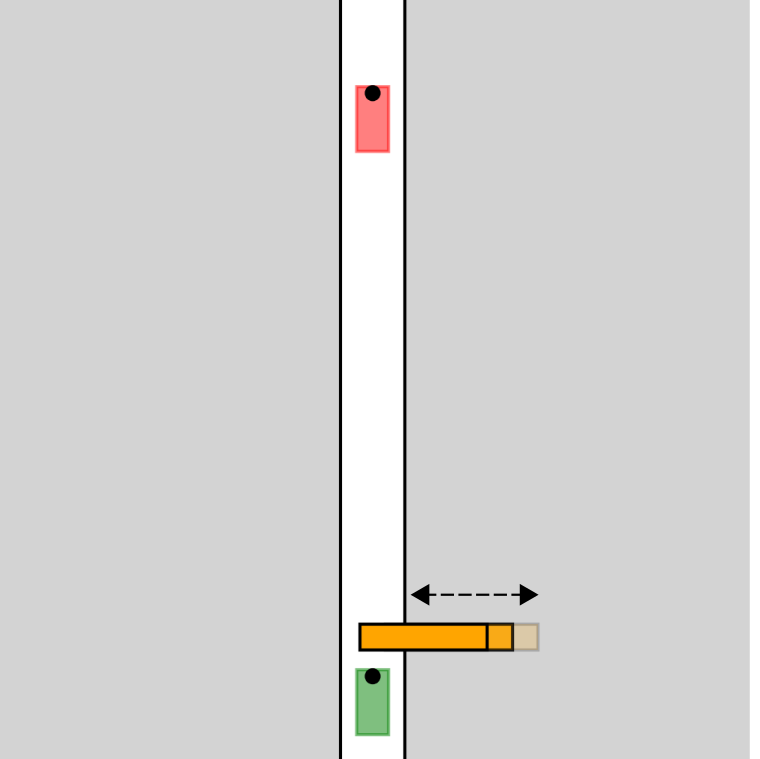}}}%
\end{minipage}\\[0.5em]

% Row 2
\begin{minipage}[t]{0.13\linewidth}\centering
  \subfloat[Elevator]{\fbox{\includegraphics[width=\linewidth]{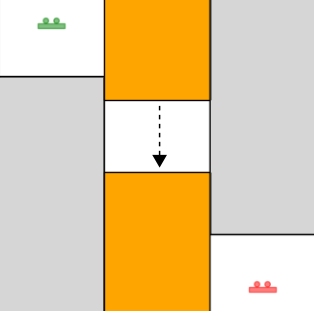}}}%
\end{minipage}\hfill
\begin{minipage}[t]{0.13\linewidth}\centering
  \subfloat[Lane Switch]{\fbox{\includegraphics[width=\linewidth]{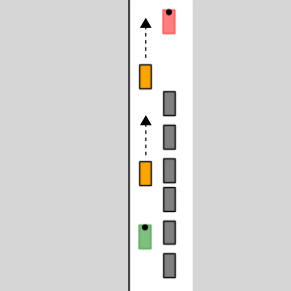}}}%
\end{minipage}\hfill
\begin{minipage}[t]{0.13\linewidth}\centering
  \subfloat[Park]{\fbox{\includegraphics[width=\linewidth]{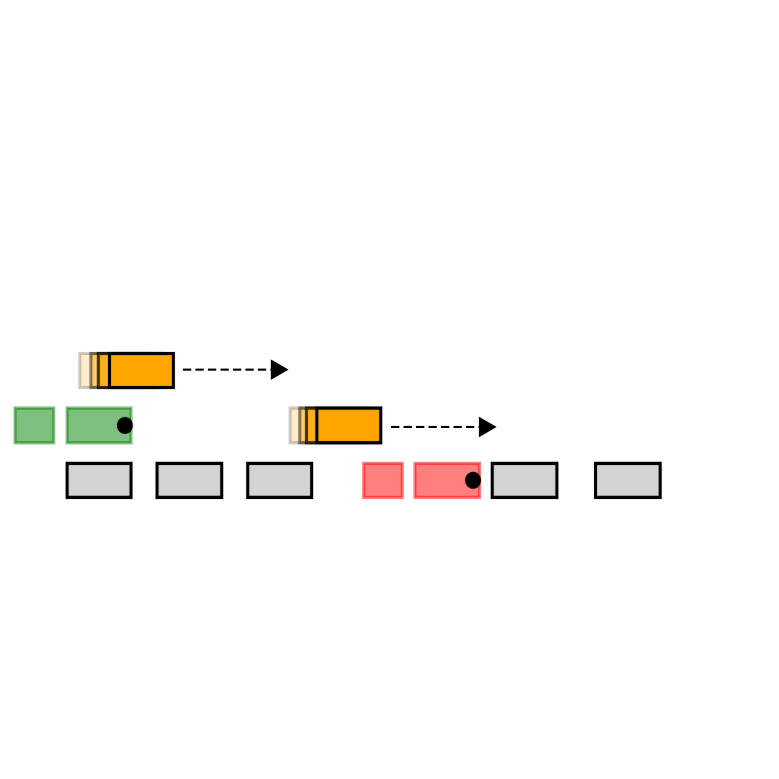}}}%
\end{minipage}\hfill
\begin{minipage}[t]{0.13\linewidth}\centering
  \subfloat[Parking Lot]{\fbox{\includegraphics[width=\linewidth]{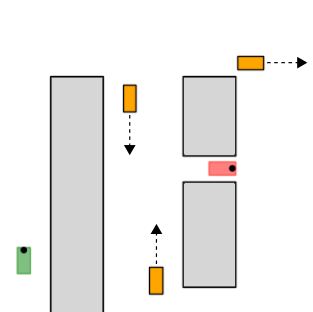}}}%
\end{minipage}\hfill
\begin{minipage}[t]{0.13\linewidth}\centering
  \subfloat[Moving Ceiling]{\fbox{\includegraphics[width=\linewidth]{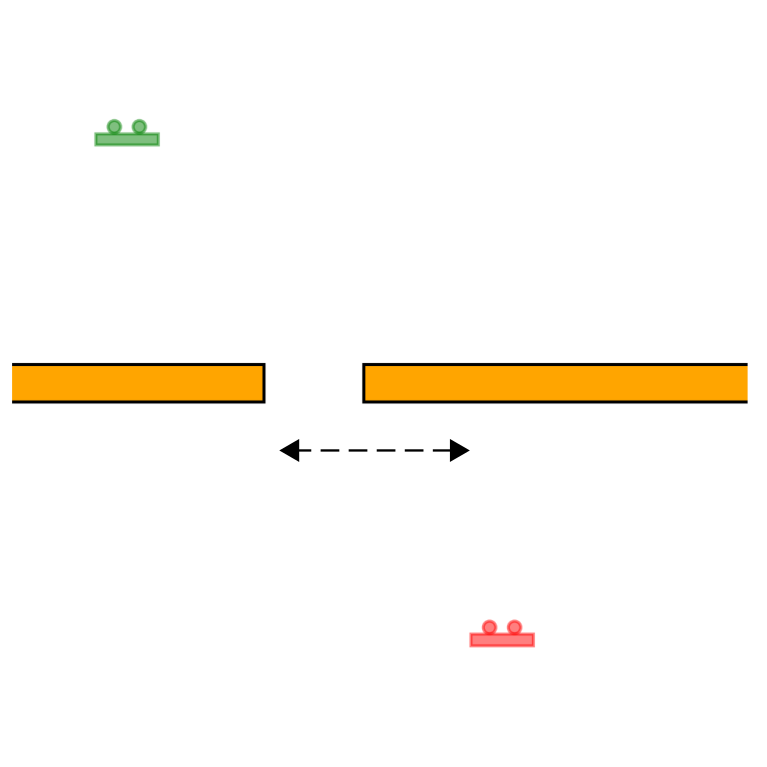}}}%
\end{minipage}\hfill
\begin{minipage}[t]{0.13\linewidth}\centering
  \subfloat[Moving Small Ceiling]{\fbox{\includegraphics[width=\linewidth]{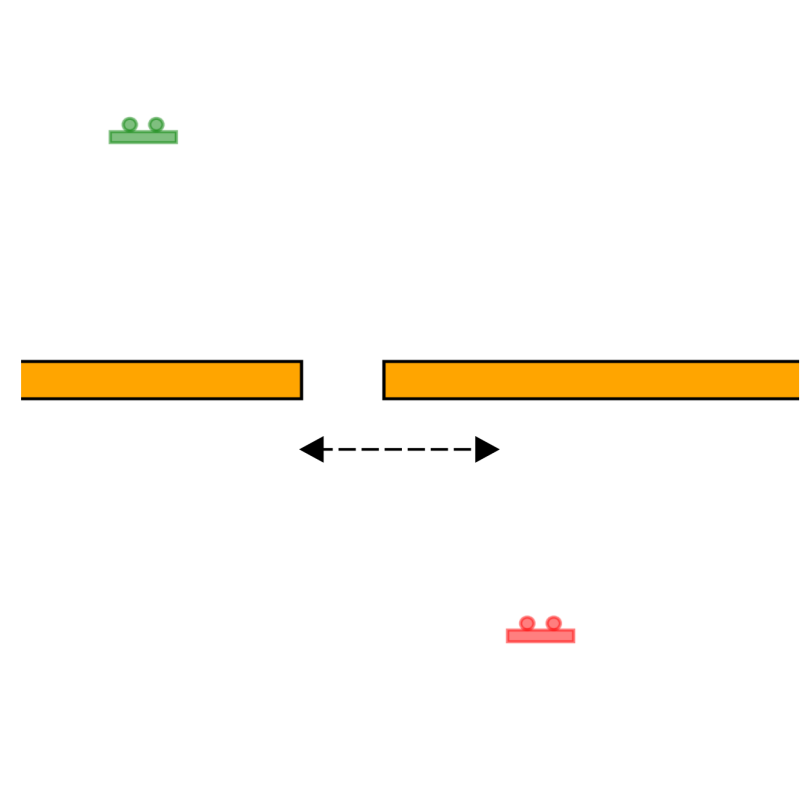}}}%
\end{minipage}\hfill
\begin{minipage}[t]{0.13\linewidth}\centering
  \subfloat[Double Ceiling]{\fbox{\includegraphics[width=\linewidth]{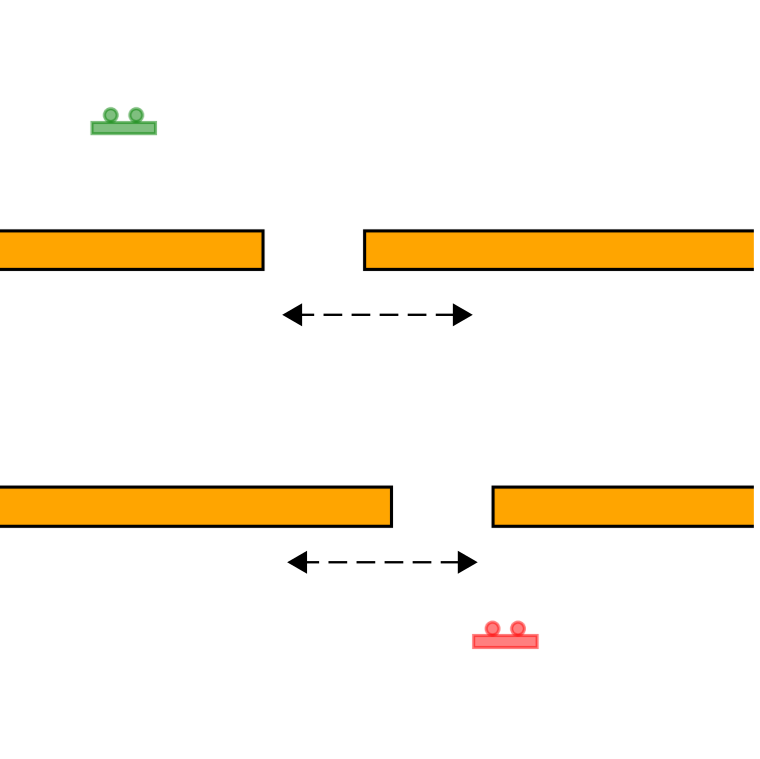}}}%
\end{minipage}

\caption{Environments used in \benchmarkName. The robot is shown in the start configuration (green) and the goal configuration (red). Static obstacles (grey) and dynamic obstacles (orange) are shown.}
\label{fig:scenarios}
\end{figure*}
% Full four-planner benchmark (new run).
% Bold = best value per metric (lowest cost / time-to-first, highest success rate; ties both bold).
\begin{table*}[t]
\caption[Planner Comparison: Db-RRT vs ST-Db-RRT vs KIST vs ST-GBRRT]{Comparing all four planners on all environments.
Bold values mark the best result per metric among the three space-time planners (lowest cost and time-to-first solution, highest success rate);
Naive Db-RRT is also bolded when it matches or beats ST-Db-RRT on that metric.
Dashes indicate that no solution was found. Grey cells mark Naive Db-RRT, which is not probabilistically complete.}
\label{tab:benchmark}
\centering
% Color for the Naive Db-RRT block — change this to whatever you like.
\definecolor{dbrrtcolor}{gray}{0.9}
% Small numeric columns: width controlled here in one place.
\newcolumntype{Y}{>{\raggedleft\arraybackslash}p{5mm}}
% Shaded version of Y, used for the Naive Db-RRT columns only.
\newcolumntype{M}{>{\columncolor{dbrrtcolor}\raggedleft\arraybackslash}p{5mm}}
% Narrow column for the "#" index.
\newcolumntype{N}{>{\raggedleft\arraybackslash}p{0.4cm}}
\newcolumntype{Z}{>{\raggedright\arraybackslash}p{2.6cm}}
% Flush booktabs rules to the vertical planner separators (no (lr) trim, no extra rule padding).
\setlength{\aboverulesep}{0pt}
\setlength{\belowrulesep}{0pt}
\setlength{\cmidrulekern}{0pt}
\renewcommand{\arraystretch}{1.15}
\begin{tabularx}{\textwidth}{@{}N X Z|MMM|YYY|YYY|YYY|}
\toprule
& & &
\multicolumn{3}{>{\columncolor{dbrrtcolor}}c|}{Naive Db-RRT} &
\multicolumn{3}{c|}{ST-Db-RRT} &
\multicolumn{3}{c|}{KIST} &
\multicolumn{3}{c|}{ST-GBRRT} \\
\cmidrule{4-6}\cmidrule{7-9}\cmidrule{10-12}\cmidrule{13-15}
\# & Scenario & Dynamics &
c [s] & $t_1$ [s] & sr &
c [s] & $t_1$ [s] & sr &
c [s] & $t_1$ [s] & sr &
c [s] & $t_1$ [s] & sr \\
\midrule
01 & Empty                 & 1st Order Unicycle & 15.5 & \textbf{0.1} & \textbf{1.0} & 14.7 & \textbf{0.5} & \textbf{1.0} & 15.1 & 1.1 & \textbf{1.0} & \textbf{14.4} & 1.9 & \textbf{1.0} \\
02 & Revolving Door        & 1st Order Unicycle & 11.5 & \textbf{0.2} & \textbf{1.0} & 11.5 & \textbf{1.1} & \textbf{1.0} & 10.7 & 1.7 & \textbf{1.0} & \textbf{10.5} & 2.6 & \textbf{1.0} \\
03 & Double Revolving Door & 1st Order Unicycle & 13.8 & \textbf{0.3} & \textbf{1.0} & 13.4 & \textbf{1.7} & \textbf{1.0} & 12.8 & 2.8 & \textbf{1.0} & \textbf{12.6} & 5.7 & \textbf{1.0} \\
04 & Multi Revolving Door  & 1st Order Unicycle & \textbf{17.2} & \textbf{0.5} & \textbf{1.0} & 17.7 & \textbf{2.7} & \textbf{1.0} & 18.4 & 5.5 & \textbf{1.0} & \textbf{16.2} & 10.6 & \textbf{1.0} \\
05 & Double Moving Windows & 1st Order Unicycle & 11.5 & \textbf{0.2} & \textbf{1.0} & 11.6 & 2.4 & \textbf{1.0} & 11.6 & \textbf{2.3} & \textbf{1.0} & \textbf{10.6} & 4.4 & \textbf{1.0} \\
06 & Shortcut              & 1st Order Unicycle & 15.2 & \textbf{0.1} & \textbf{1.0} & 14.7 & \textbf{0.8} & \textbf{1.0} & 14.6 & 2.1 & \textbf{1.0} & \textbf{14.1} & 3.4 & \textbf{1.0} \\
07 & Bugtrap               & 1st Order Unicycle & \textbf{28.8} & \textbf{0.6} & \textbf{1.0} & \textbf{32.2} & \textbf{32.3} & \textbf{1.0} & 41.1 & 42.3 & \textbf{1.0} & -- & -- & 0.0 \\
08 & Wait                  & 1st Order Unicycle & -- & -- & 0.0 & 39.4 & \textbf{2.0} & \textbf{1.0} & \textbf{39.1} & 32.2 & 0.7 & -- & -- & 0.0 \\
09 & Elevator              & 1st Order Unicycle & 17.9 & 35.4 & 0.9 & 17.4 & 18.0 & 0.5 & 18.4 & 16.9 & 0.6 & \textbf{16.6} & \textbf{15.4} & \textbf{1.0} \\
10 & Empty                 & 2nd Order Unicycle & 21.7 & \textbf{0.2} & \textbf{1.0} & 19.6 & \textbf{0.5} & \textbf{1.0} & 19.9 & 1.3 & \textbf{1.0} & \textbf{18.7} & 44.8 & 0.9 \\
11 & Revolving Door        & 2nd Order Unicycle & 13.4 & \textbf{0.2} & \textbf{1.0} & 12.7 & \textbf{1.2} & \textbf{1.0} & 13.2 & 3.8 & \textbf{1.0} & \textbf{12.2} & 10.7 & \textbf{1.0} \\
12 & Double Revolving Door & 2nd Order Unicycle & 22.4 & \textbf{0.3} & 0.8 & \textbf{20.7} & \textbf{1.9} & \textbf{1.0} & 23.3 & 5.4 & \textbf{1.0} & -- & -- & 0.0 \\
13 & Multi Revolving Door  & 2nd Order Unicycle & 23.6 & \textbf{0.5} & \textbf{1.0} & 22.5 & \textbf{2.5} & \textbf{1.0} & 34.3 & 7.7 & \textbf{1.0} & \textbf{21.5} & 79.9 & 0.8 \\
14 & Double Moving Windows & 2nd Order Unicycle & 15.9 & \textbf{0.3} & \textbf{1.0} & 17.7 & \textbf{3.5} & \textbf{1.0} & 27.6 & 8.5 & \textbf{1.0} & \textbf{15.2} & 74.5 & 0.2 \\
15 & Wait                  & 2nd Order Unicycle & -- & -- & 0.0 & \textbf{41.9} & \textbf{38.5} & \textbf{0.7} & -- & -- & 0.0 & -- & -- & 0.0 \\
16 & Lane Switch           & 2nd Order Unicycle & 30.6 & \textbf{0.3} & \textbf{1.0} & \textbf{29.8} & \textbf{3.1} & \textbf{1.0} & 29.9 & 7.0 & \textbf{1.0} & -- & -- & 0.0 \\
17 & Empty                 &  Car with Trailer      & 14.4 & \textbf{0.1} & \textbf{1.0} & 17.9 & 1.3 & \textbf{1.0} & 14.1 & \textbf{0.5} & \textbf{1.0} & \textbf{13.2} & 4.3 & \textbf{1.0} \\
18 & Lane Switch           &  Car with Trailer      & 29.4 & \textbf{0.3} & \textbf{1.0} & 31.6 & \textbf{2.1} & \textbf{1.0} & \textbf{28.7} & 26.9 & 0.9 & -- & -- & 0.0 \\
19 & Park                  &  Car with Trailer      & 7.8 & 0.3 & \textbf{1.0} & 8.2 & \textbf{0.4} & \textbf{1.0} & 7.7 & 0.9 & \textbf{1.0} & \textbf{7.0} & 1.4 & \textbf{1.0} \\
20 & Parking Lot           & Car with Trailer     & \textbf{25.1} & \textbf{0.4} & \textbf{1.0} & 30.0 & \textbf{2.8} & \textbf{1.0} & \textbf{25.8} & 16.8 & \textbf{1.0} & -- & -- & 0.0 \\
21 & Empty                 & Planar Rotor   & 2.8 & \textbf{0.6} & \textbf{1.0} & 5.8 & 11.0 & \textbf{1.0} & 3.5 & \textbf{4.1} & \textbf{1.0} & \textbf{2.6} & 94.4 & 0.5 \\
22 & Moving Ceiling        & Planar Rotor   & 5.3 & \textbf{4.7} & \textbf{1.0} & 7.6 & \textbf{18.0} & \textbf{1.0} & \textbf{5.9} & 19.7 & \textbf{1.0} & -- & -- & 0.0 \\
23 & Moving Small Ceiling  & Planar Rotor   & 7.1 & 4.1 & \textbf{1.0} & 9.5 & 33.9 & \textbf{1.0} & \textbf{6.8} & \textbf{28.0} & \textbf{1.0} & -- & -- & 0.0 \\
24 & Double Ceiling        & Planar Rotor   & 8.3 & 6.1 & 0.9 & 12.2 & 73.7 & 0.9 & \textbf{7.6} & \textbf{40.0} & \textbf{1.0} & -- & -- & 0.0 \\
25 & Elevator              & Planar Rotor   & 15.6 & \textbf{5.9} & \textbf{0.8} & 19.7 & \textbf{80.8} & 0.6 & \textbf{16.9} & 91.2 & \textbf{0.8} & -- & -- & 0.0 \\
\bottomrule
\end{tabularx}

\end{table*}

\section{Benchmark Results}

This section evaluates Naive Db-RRT and the three space-time planners on the \benchmarkName set. The
algorithms are implemented in
\textit{Dynoplan}~\cite{ortizharo2024idbrrtsamplingbasedkinodynamicmotion},
which provides the dynamical systems including motion primitives and
infrastructure.
For this work, we extended the existing infrastructure to deal with dynamic obstacles.

The evaluation is presented in three parts. First, we present a hyperparameter study
to select a common
parameter set. Second, we compare Db-RRT and ST-Db-RRT to show the influence of the time dimension. 
Third, we compare ST-Db-RRT, ST-GBRRT,
and KIST in terms of success rate, time to first solution, and solution cost.

\paragraph{Hardware and Benchmark Parameters}

Each individual run of a planner uses a single core of an AMD 
%Ryzen Threadripper PRO 5975WX 
CPU running at 3.60 GHz with 128 GB RAM using Ubuntu 22.04.
Each run has a time limit of $180$\,s and returns a solution cost $c$, the time to the first solution $t_1$, and the success rate $sr$ for each
algorithm. The results are given as
averages over a set of $N=20$ runs.
%For each evaluation, the best value per metric is highlighted in bold.

\paragraph{Environments}
We use a suite of 15 environments comprising 25 instances across four dynamical systems, including 10 newly designed layouts and 4 dynamic adaptations of static environments from \textit{Dynobench} \cite{ortizharo2023idbaiterativesearchoptimization} (Fig.~\ref{fig:scenarios} shows the 14 dynamic layouts; Empty is omitted).
%An overview of the environments is given in Table \ref{tab:environments}, grouped by the dynamics model they target.

%Beyond their spatial layout, the environments differ in their \emph{temporal structure}, that is, when a passage becomes traversable and for how long it stays open. Most impose only a loose timing requirement, where the goal is reachable across a wide range of arrival times, whereas the \textit{Wait} and \textit{Elevator} admit a solution only within a narrow window at a specific moment within a confined space. This temporal dimension is precisely what the space-time sampling schedule must resolve (Section~\ref{sec:bench_timebounds}), so the relevant timing of each environment is highlighted in Table \ref{tab:environments} and a detailed description of the environments follows in the next sections.

\paragraph{Dynamical Systems}

In the benchmarks, we use four dynamical systems. The first two are a 1st and 2nd order unicycle system. The third is a car-with-trailer model, which reflects a
realistic driving system. Finally, we use a planar rotor as an underactuated flying system~\cite{ortizharo2024idbrrtsamplingbasedkinodynamicmotion}.
%The \textit{Lane Switch} environment
%(Figure~\ref{fig:car_envs}, top) simulates a
%lane-change maneuver in crowded traffic. The \textit{Park} environment is a
%parallel-parking scenario in which another car initially blocks the parking
%space. The \textit{Parking Lot} environment (Figure~\ref{fig:car_envs}, bottom) requires the car to traverse a car
%park into a parking spot while avoiding other moving cars.

%The Planar Rotor environments require the rotor to fly through narrow gaps that translate back and forth. The \textit{Moving Window} environment  is the simplest and \textit{Double Moving Window} (bottom) and \textit{Moving Small Window} tighten this timing further with two oppositely moving gaps and a narrower gap, respectively. 
%As with the unicycle, the \textit{Elevator} environment requires the rotor to move tightly alongside other obstacles to avoid collision with a late feasible arrival time.

%%%%%%%%%%%%%%%%%%%%%%%%%%%%%%%%%%%%%%%%%%%%%%%%%%%%%%%%%%%%%%%%%%%%%%%%%%%%%%%%
\subsection{Planner Parameter Study}
%%%%%%%%%%%%%%%%%%%%%%%%%%%%%%%%%%%%%%%%%%%%%%%%%%%%%%%%%%%%%%%%%%%%%%%%%%%%%%%%

%groups. Group 1 contains shared static
%parameters, including
%goal region
%($0.3$), distance metric, and motion primitives.
%Group 2 are planner specific parameters, involving, for KIST, the maximal queue
%size of the \textit{Heuristic Completion} policy using $|Q|=500$. 
%For ST-GBRRT, the maximal heuristic radius is set to $r_{\mathrm{max}}=0.3$.
%Finally Group 3 contains shared parameters, which affect the planner
%performance and should depend on the individual planner. For this group, we
%conduct a hyperparameter optimization using \benchmarkName to ensure each
%planner uses the best individual parameter.

To choose the best set of parameters for ST-Db-RRT, ST-GBRRT, and KIST, we use \benchmarkName to conduct a hyperparameter optimization on four parameters.
First, we compare time-optimal versus random motion primitives. Our results
indicate that time-optimal primitives slightly outperform random primitives.
Second, we investigate the motion primitive density and discontinuity bound.
From the results, we have chosen a low to medium density of $1000$ primitives ($3000$
for planar rotor) and a discontinuity bound of $\delta=0.3$. 
Third, we compared the sampling time bounds, including the batch size to sample
and the relative increase of the upper bound. Our analysis gives a value of
$b=1200$ and $f=1.7$ as optimal values for this dataset.
Finally, we optimized for the time weight in the distance function, finding that a small value of
$\lambda=0.2$ gives the best results.

\subsection{Planner Comparisons}

Using the optimized parameter set, we compare all three space-time planners and Naive Db-RRT on the scenario set. The results are depicted in Tab.~\ref{tab:benchmark}. 

%%%%%%%%%%%%%%%%%%%%%%%%%%%%%%%%%%%%%%%%%%%%%%%%%%%%%%%%%%%%%%%%%%%%%%%%%%%%%%%%
\subsubsection{Naive Db-RRT vs ST-Db-RRT\label{sec:bench_db_stdb}}
%%%%%%%%%%%%%%%%%%%%%%%%%%%%%%%%%%%%%%%%%%%%%%%%%%%%%%%%%%%%%%%%%%%%%%%%%%%%%%%%
%We study the influence of space-time planning by comparing the naive Db-RRT~\cite{ortizharo2024idbrrtsamplingbasedkinodynamicmotion} (which is not PC) with its space-time variant ST-Db-RRT. Since ST-Db-RRT is a bidirectional planner, we run Db-RRT also in its bidirectional variant with the default parameter of $p_{\mathrm{forward}}=0.5$ probability to expand the forward tree. Both planners use the same amount of motion primitives, discontinuity bound, and the same fixed-time trajectory optimization.

Db-RRT reaches a success rate of $sr=1.0$ on 19 of the 25 instances and returns a first solution within $t_1=0.6$\,s on 17 of the 18 unicycle and car instances it solves, up to $54\times$ faster than ST-Db-RRT in the \textit{Bugtrap} environment ($t_1=0.6$\,s against $t_1=32.3$\,s).

However, in the \textit{Wait} environments, which require the robot to wait for a long time in a confined space, Db-RRT fails while ST-Db-RRT solves them with $sr=1.0$ for the 1st Order Unicycle and $sr=0.7$ for the 2nd Order Unicycle. Outside the \textit{Wait} environments the success rates of the two planners are close, and the time dimension is not by itself an advantage: on all five Planar Rotor instances Db-RRT matches or exceeds ST-Db-RRT.

Planning in space-time does not pay off as clearly in the solution cost. ST-Db-RRT is cheaper on only 9 of the 23 jointly solved instances, and all nine are unicycle instances, where the margin is a moderate $3\%$ to $10\%$. On the Car with Trailer and Planar Rotor systems the ordering inverts: Db-RRT attains the lower cost on every one of these nine instances, by $26\%$ to $107\%$ on the rotor, e.g. in the \textit{Double Ceiling} ($c=8.3$\,s against $c=12.2$\,s).

%The benefit of planning in space-time is therefore feasibility under tight temporal constraints rather than cost, paid for with an order of magnitude in time to first solution.

%%%%%%%%%%%%%%%%%%%%%%%%%%%%%%%%%%%%%%%%%%%%%%%%%%%%%%%%%%%%%%%%%%%%%%%%%%%%%%%%
\subsubsection{Space-Time Planner Comparison}
%%%%%%%%%%%%%%%%%%%%%%%%%%%%%%%%%%%%%%%%%%%%%%%%%%%%%%%%%%%%%%%%%%%%%%%%%%%%%%%%

We compare the space-time planners in terms of success rate, time to first solution, and solution cost.
%When comparing all space-time planners, we observe that ST-Db-RRT is generally the strongest planner in this comparison. It solves every instance, reaches first solutions within seconds, and its final solution cost stays close to the best observed value on every instance. 
%The strengths of the two optimization-free planners are more localized: KIST solves most of the instances but consistently slower, while ST-GBRRT converges to the lowest costs on the simpler instances but fails on large parts of the benchmark.
%This pattern is visible in the cost-success plots (Fig.~\ref{fig:cost-success1}): ST-Db-RRT reaches its first solution fastest, while ST-GBRRT converges to slightly lower costs. On the higher-dimensional 2nd Order Unicycle Multi Revolving Door instance, ST-GBRRT does not reach a success rate of $100\%$.
%The cost-success plot in Fig.~\ref{fig:cost-success1} (bottom row) show two cases where ST-Db-RRT doesn't reach $100\%$ success rate. 
%In the bugtrap environment, we can see that if ST-Db-RRT finds a solution, it
%finds it very quickly, but only does so in $<80\%$ of the runs. In the Elevator environment, ST-Db-RRT finds a solution in only $30\%$ of the runs, while KIST gets up to $60\%$ and ST-GBRRT even up to $100\%$.

\paragraph{Success Rate}
ST-Db-RRT achieves $100\%$ on 21 of the 25 instances and is the only planner that solves every instance at least once. Most notably, it is the only planner to find any solution in the 2nd Order Unicycle \textit{Wait} instance ($sr=0.7$), which forces the system to wait roughly \SI{30}{\second} in a confined region. Its success rate is surpassed on only three instances, all with a narrow arrival window: the 1st Order Unicycle \textit{Elevator} ($sr=0.5$ against $sr=1.0$ for ST-GBRRT), the Planar Rotor \textit{Double Ceiling} ($sr=0.9$ against $sr=1.0$ for KIST) and the Planar Rotor \textit{Elevator} ($sr=0.6$ against $sr=0.8$), confirming the recurring weakness of ST-Db-RRT in those environments.
%already observed in the parameter benchmarks.
KIST is similarly reliable ($sr=1.0$ on 20 of the 25 instances), and its success rate declines only on the temporally hardest problems: it drops to $sr=0.7$ in the 1st Order Unicycle \textit{Wait}, to $sr=0.6$ in the 1st Order Unicycle \textit{Elevator}, and fails the 2nd Order Unicycle \textit{Wait}. On the Planar Rotor it is the most reliable planner of the three, reaching $sr=1.0$ on four of the five instances. ST-GBRRT has the most limited coverage and fails on 11 of the 25 instances, mostly on the long-horizon problems \textit{Bugtrap}, \textit{Wait}, \textit{Lane Switch} and \textit{Parking Lot}, and on the Planar Rotor, where only the \textit{Empty} instance is solved. On the 2nd Order Unicycle it additionally fails in three scenarios outright and drops to $sr=0.2$ in the \textit{Double Moving Windows}.

\paragraph{Time to First Solution}
The clearest separation between the planners appears in the time to the first solution. ST-Db-RRT is the fastest of the three planners on 19 of the 25 instances and stays below $t_1=4$\,s on 17 of the 20 unicycle and car instances, the exceptions being \textit{Bugtrap}, \textit{Wait}, and the 1st Order Unicycle \textit{Elevator}. Its $t_1$ is thereby largely insensitive to the temporal constraints that slow the optimization-free planners down by an order of magnitude: KIST needs $t_1=32.2$\,s against $t_1=2.0$\,s in the 1st Order Unicycle \textit{Wait}, a $16\times$ speed-up for ST-Db-RRT. On the Planar Rotor, however, this ordering no longer holds: KIST reaches a first solution faster than ST-Db-RRT on three of the five instances, clearly so in the \textit{Double Ceiling} ($t_1=40.0$\,s against $t_1=73.7$\,s). ST-GBRRT is the slowest planner on almost every instance it solves: on the 2nd Order Unicycle it needs up $32\times$ longer than ST-Db-RRT, e.g. $t_1=79.9$\,s against $t_1=2.5$\,s in the \textit{Multi Revolving Door}, which also explains its reduced success rates there.

\paragraph{Solution Cost}
%The solution cost separates the planners differently from the two previous metrics. R
ST-GBRRT converges to the lowest cost on every one of the 14 instances it solves, despite reaching its first solution last. KIST attains the best cost on 7 instances, and is the cheapest planner on four of the five Planar Rotor instances. Its cost inflates, however, in the dynamically crowded 2nd Order Unicycle environments, where it exceeds ST-Db-RRT by roughly $50\%$ in the \textit{Multi Revolving Door} %($c=34.3$\,s against $c=22.5$\,s)%
and the \textit{Double Moving Windows}.%($c=27.6$\,s against $c=17.7$\,s)%. ST-Db-RRT holds the best cost on 4 instances.
%, and on the 2nd Order Unicycle the best cost is always attained by either ST-Db-RRT or ST-GBRRT.
ST-Db-RRT, however, is no longer uniformly close to the best value: its cost stays within $10\%$ of the best observed value on all but one unicycle instance, but it is $36\%$ above the best cost in the Car with Trailer \textit{Empty} and between $29\%$ and $123\%$ above it on the Planar Rotor.

\section{Conclusion}

We presented \benchmarkName, a benchmark environment to evaluate kinodynamic space-time motion planners in dynamic environments. To demonstrate its utility and establish strong baselines, we developed three dedicated planners that fuse kinodynamic and space-time methods: We used state-of-the-art kinodynamic planners iDb-RRT~\cite{ortizharo2024idbrrtsamplingbasedkinodynamicmotion} and GBRRT~\cite{gbrrt} and combined them with aspects of the space-time planner ST-RRT*~\cite{strrt}. The results are new planners ST-Db-RRT and ST-GBRRT, respectively. Additionally, we combined different aspects of those planners leading to the KIST planner. For those three planners, we conducted a theoretical analysis of completeness properties, and then used \benchmarkName to conduct (a) a hyperparameter analysis, (b) a comparison between iDb-RRT and ST-Db-RRT, and (c) a full evaluation of ST-Db-RRT, ST-GBRRT, and KIST on a benchmark set of 25 scenarios involving dynamic obstacles and four different dynamical systems.

Our evaluation showed that the three planners differ in all three metrics.
ST-Db-RRT remains resolution-complete and is the only planner to solve every dynamic instance while it reaches the first solution up to $90\times$ faster than the optimization-free planners, demonstrating that deferring feasibility to a single repair step wins coverage and runtime under a fixed time budget. It does not, however, win on cost: ST-GBRRT converges to the lowest cost on every instance its slower search can cover, while KIST is both the most reliable and the cheapest planner on the Planar Rotor. This picture inverts on the higher-dimensional systems and wherever the feasible arrival window is narrow, as in the \textit{Elevator} and moving-ceiling instances, revealing that optimization-free planning remains valuable when the optimization struggles.

%The theoretical analysis showed that the completeness properties of the %underlying planners are not preserved automatically: a naive application of %Db-RRT to dynamic environments is incomplete, which the benchmarks confirm in %the \textit{Wait} environment. Lifting the planners into space-time restores completeness, with ST-Db-RRT being resolution complete once a time bound is given, while KIST and ST-GBRRT retain probabilistic completeness. In static environments, however, the additional dimension delays the first solution by roughly an order of magnitude, making space-time planning a more expensive variant to be enabled when the temporal structure of the environment demands it.
%Our hyperparameter evaluation showed that all three planners are sensitive to their central parameters. The motion primitive set should be kept as coarse as the environments permit, a small time weight in the distance metric is the robust default, and the best time bound schedule depends on where the feasible arrival window lies and how narrow it is.

There are multiple immediate extensions of the planners. As outlined, a more informed arrival-time estimate could reach late feasible arrival windows more reliably, kinodynamic rewiring in space-time could recover the asymptotic optimality of ST-RRT*, and variable-time trajectory optimization could further improve solution quality. 
% However, we believe that by developing \benchmarkName and the novel planners ST-Db-RRT, ST-GBRRT, and KIST, we have created the first sampling-based planners for general kinodynamic systems that plan through space-time without requiring a steering function or a known arrival time. We have thereby taken a significant step towards general-purpose kinodynamic planning in dynamic environments.

%\appendix
%\input{src/appendix.tex}

\bibliographystyle{IEEEtran}
\balance
\bibliography{bib/general}
\end{document}